\documentclass[10pt,twocolumn,letterpaper]{article}

\usepackage{iccv}
\usepackage{times}
\usepackage{epsfig}
\usepackage{graphicx}
\usepackage{amsmath}
\usepackage{amssymb}
\usepackage{times}
\usepackage{epsfig}
\usepackage{graphicx}
\usepackage{amsmath}
\usepackage{amssymb}
\usepackage{enumitem}
\usepackage{bbding}
\usepackage{pifont}
\usepackage{wasysym}
\usepackage{color}
\usepackage{multirow}
\usepackage[switch]{lineno}
\usepackage{booktabs}
\usepackage{subcaption}
\usepackage{authblk}
\newcommand{\kwang}[1]{{\color{black}{#1}}}
\newcommand{\pxj}[1]{{\color{black}{#1}}}

\usepackage[pagebackref=true,breaklinks=true,letterpaper=true,colorlinks,bookmarks=false]{hyperref}

\iccvfinalcopy 

\def\iccvPaperID{4098} 

\ificcvfinal\fi

\begin{document}

\title{AU-Guided Unsupervised Domain Adaptive Facial Expression Recognition}


\author[1]{Kai Wang$^*$}
\author[1]{Yuxin Gu\thanks{Equally-contributed first authors}}
\author[2]{Xiaojiang Peng$^*$}

\author[3]{Panpan Zhang}
\author[1]{Baigui Sun}
\author[1]{Hao Li \thanks{Corresponding author (lihao.lh@alibaba-inc.com)}}
\affil[1]{Alibaba DAMO Acadmey}
\affil[2]{Shenzhen Technology University, China}
\affil[3]{Fudan University, China}







\maketitle
\ificcvfinal\thispagestyle{empty}\fi

\begin{abstract}
The domain diversities including inconsistent annotation and varied image collection conditions inevitably exist among different facial expression recognition (FER) datasets, which pose an evident challenge for adapting the FER model trained on one dataset to another one. Recent works mainly focus on domain-invariant deep feature learning with adversarial learning mechanism, ignoring the sibling facial action unit (AU) detection task which has obtained great progress. Considering AUs objectively determine facial expressions, this paper proposes an AU-guided unsupervised Domain Adaptive FER (AdaFER) framework to relieve the annotation bias between different FER datasets. In AdaFER, we first leverage an advanced model for AU detection on both source and target domain. Then, we compare the AU results to perform AU-guided annotating, i.e., target faces that own the same AUs with source faces would inherit the labels from source domain. Meanwhile, to achieve domain-invariant compact features, we utilize an AU-guided triplet training which randomly collects anchor-positive-negative triplets on both domains with AUs.
We conduct extensive experiments on several popular benchmarks and show that AdaFER achieves state-of-the-art results on all these benchmarks.
\end{abstract}

\section{Introduction}
\label{intro}
Facial expression is one of the most important modalities in human emotional communication. Accurately recognizing facial expressions helps understand various human emotions and intents, which is applied in wide-range applications, such as human-computer interaction \cite{cowie2001emotion}, service robots \cite{giorgana2011facial}, and medicine treatment \cite{jiang2012brain}. Both in industry and academy areas, in the past decades, many well-labeled datasets \cite{lucey2010extended,valstar2010induced,zhao2011facial,dhall2011static,barsoum2016training,mollahosseini2017affectnet,fabian2016emotionet,li2017reliable} and high-performance algorithms \cite{wang2020region,li2018occlusion,wang2020suppressing} have been proposed to automatically recognize facial expressions in the past decades.

\begin{figure}[tp]
\centering
	\includegraphics[width=0.5\textwidth]{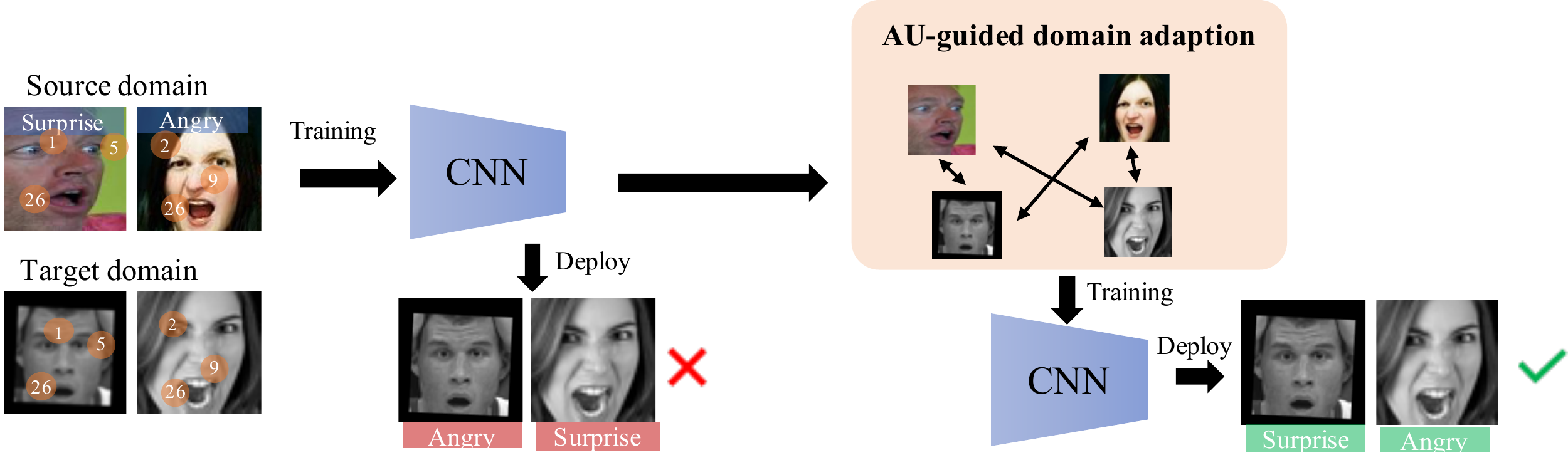}
	\caption{Illustration of unsupervised domain adaptive facial expression recognition. A model trained on source domain usually fails on target domain due to subjective inconsistent annotations and different imaging conditions. AdaFER leverages the objective facial action units for an auxiliary training, effectively relieve the annotation bias between source and target domains.}
	
	\label{fig:motivation}
\end{figure}

In general, deep learning based facial expression recognition (FER) methods~\cite{mollahosseini2017affectnet,fabian2016emotionet,li2017reliable,wang2020region,li2018occlusion,wang2020suppressing} achieve high performance only when the training domain is identical or similar to the testing domain. However, due to the diversities of inconsistent annotation and different image collection condition, there inevitably exists annotation bias (domain gap) among different datasests, which poses an evident challenge for adapting the FER model trained on one dataset to another one.
As shown in Figure \ref{fig:motivation}, a naive cross-domain method that deploying the trained model on different target domain often fails. To this end, many methods have been presented to mitigate the domain shift in FER~\cite{yan2019cross,xie2020adversarial,zong2018domain}, while almost all the methods follow general domain adaption algorithms and ignore the sibling facial action unit (AU) detection task. 
\kwang{Action Units (AUs) represent the movement of the facial muscles, which have lower bias than subjective facial expression annotations. Intuitively, AUs can be regarded as auxiliary cues to alleviate the annotation bias \pxj{and data} bias (domain gap) among different FER datasets.}


In this paper, considering the remarkable progress and stable performance of AU detection~\cite{mlcr,ji2020multiple}, we propose an AU-guided unsupervised Domain Adaptive FER (AdaFER). The AdaFER consists of two crucial modules: AU-Guided  Annotating (AGA) and AU-Guided Triplet Training (AGT). Given two groups of images from source domain and target domain, we first utilize an advanced pretrained AU detection model to extract AUs coding on both domains. Then, we perform AU-guided annotating as following. For each image on target domain, we use its AUs coding to query the source domain. All these images on source domain with the same AUs coding as the query image will be used for annotating. By default, the query image is assigned with a soft label which is the statistic label distribution of the retrieval images. Further, to achieve structure-reliable and compact facial expression features, we perform the AU-guided triplet training. Each triplet is generated by comparing the AUs coding of source and target images. For example, when a source image is used as an anchor, we randomly sample a positive(negative) sample that has same(different) AUs as the anchor from all target images. 
With AdaFER, we are able to fine-tune a source-pretrained model on target domain with both pseudo soft labels and triplet loss, which effectively prevent FER performance degradation on target domain.


Overall, our contributions can be summarised as follow,
\begin{itemize}
\item{We heuristically utilize the relationship between action units and facial expressions for cross-domain facial expression recognition, and propose AU-guided unsupervised Domain Adaptive FER (AdaFER).}

\item{We elaborately design an AU-Guided Annotating module to assign soft labels for target domain and an AU-Guided Triplet Training module to learn structure-reliable and compact facial expression features.}

\item{We conduct extensive experiments on several popular benchmarks and significantly outperform the state-of-the-art methods.}

\end{itemize}

\section{Related Work}
\label{relatedwork}
\subsection{Facial Expression Recognition}
Recently, facial expression recognition has achieved a significant progress due to the well-designed feature extraction methods and high-performance algorithms. They first detect and align faces using several popular face detectors like MTCNN \cite{zhang2016joint} and Dlib \cite{amos2016openface}. For feature extraction, a large number of methods focus on modeling the facial geometry and appearance features to help facial expression recognition. From the feature type view, these features can be generally divided into hand-craft feature and deep-learning feature. The hand-craft feature usually contains texture-based features and geometry-based features \cite{ng2003sift,shan2009facial,dalal2005histograms}. Sometimes they are used in a combination called hybrid features. 
For deep-learning feature, Tang \cite{tang2013deep} and Kahou \textit{et al.}\cite{kahou2013combining}  utilize deep CNNs for feature extraction, and win the FER2013 and Emotiw2013 challenge, respectively. Zhou \textit{et al.} \cite{zhou2019exploring} achieve a remarkable result in Emotiw2019 multi-modal emotion recognition challenge by using audio-video deep fusion method. To address the pose variant and occlusion in FER, Wang \textit{et al.} \cite{wang2020region} and Li \textit{et al.} \cite{li2018occlusion} design region-based attention network. Wang \textit{et al.} \cite{wang2020suppressing} propose Self-Cure Network to suppress uncertainty samples in FER datasets. Liu \textit{et al.} \cite{liu2015inspired} introduce a Facial Action Units (FAUs) based network for expression recognition. Daniel \textit{et al.} \cite{2890247} present a cross-platform real-time multi-face expression recognition toolkit using the (FAUs), which shows the robustness of FAUs in FER system.

\begin{figure*}[tp]
	\includegraphics[width=\textwidth]{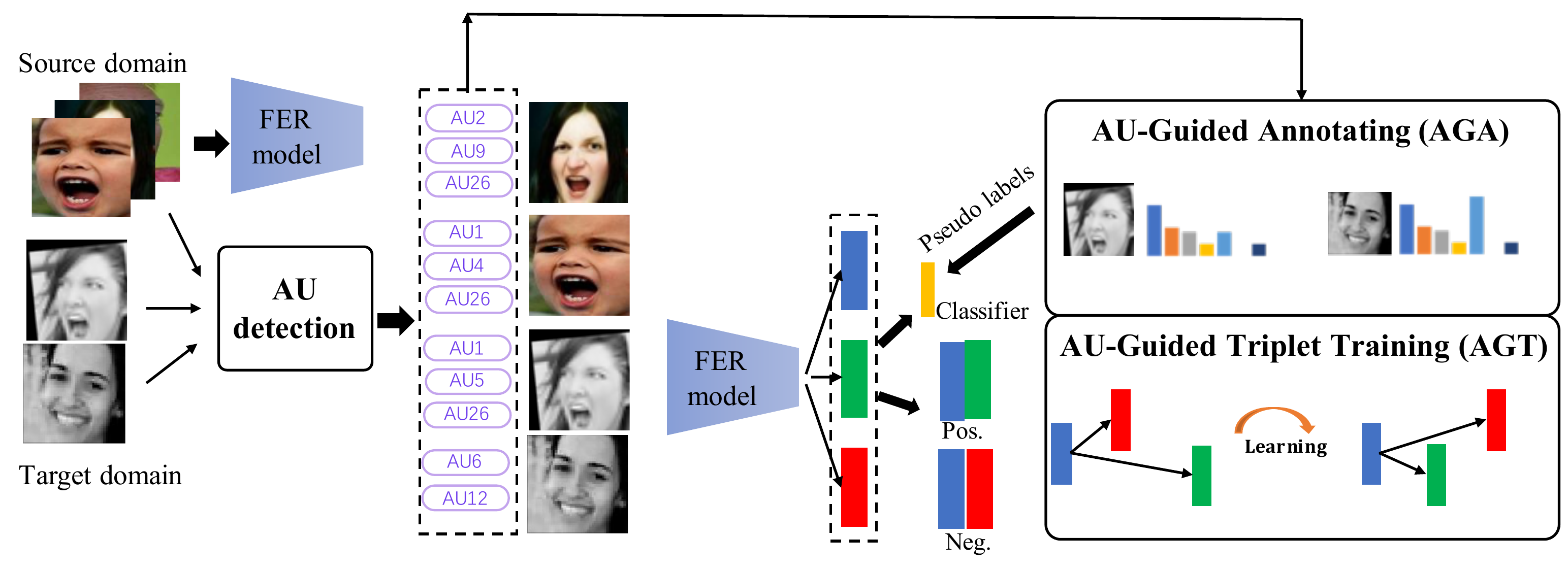}
	\caption{The pipeline of our AdaFER. We first utilize a pretrained AU detection model to extract facial action units for both source and target domain images. Then an AU-Guided Annotating (AGA) module assigns a soft/hard pseudo label for each image in target domain by comparing the AUs between source and target domain. Meanwhile, we mine triplets among source and target domains according AUs and apply Triplet loss for training. We jointly train the FER model on source domain with ground-truths and target domain with pseudo labels and triplets. AdaFER makes full use of the objective AUs to bridge the gap of FER datasets caused by subjective inconsistent annotation and different imaging condition.}
	\label{fig:pipeline}
\end{figure*}

\subsection{Cross-Domain FER}
It is inevitable to exist distribution divergences among different facial expression datasets due to the variant collecting conditions and annotating subjectiveness. In past decades, cross-domain FER (CD-FER) has got more attention \cite{chu2016selective,li2020deeper,miao2012cross,sangineto2014we,yan2016transfer,yan2016cross,yan2019cross,zheng2016cross,zhu2016discriminative,zong2018domain}. Generally, the CD-FER can be divided into semi-supervised-based, unsupervised-based, and generate-based methods. Semi-supervised-based methods \cite{levi2015emotion,yan2016transfer} apply Convolutional Neural Network (CNN) model to train a classification model using limited labeled samples from target domain. For unsupervised-based methods, Valstar \textit{et al.} \cite{6020812} use a Gabor-feature based landmarks detector to localize facial and track these points in facial sequences to model temporal facial activation for facial expression recognition. They trained the recognition model using the CK database and performed the test in the MMI database for cross-validation. Zheng \textit{et al.}\cite{7465718}  propose a transductive transfer subspace learning method using labeled source domain images and unlabelled auxiliary target domain images to jointly learn a discriminative subspace. For generate-based methods, Zong \textit{et al.} \cite{8268553}  propose a domain regeneration framework (DR) that aims at learning a domain regenerator to regenerate samples from source and target databases, respectively. Wang \textit{et al.} \cite{wang2018unsupervised} introduce an unsupervised domain adaptation method using generative adversarial network (GAN) on the target dataset and give the unlabelled GAN generated samples distributed pseudo labels dynamically according to the current prediction probabilities. In order to understand the conditional probability distributions differences between source and target datasets, Li \textit{et al.} \cite{li2020deeper} develop a deep emotion-conditional adaption network that simultaneously considers conditional distribution bias and expression class imbalance problem in CD-FER. Chen \textit{et al.} \cite{xie2020adversarial} propose a Adversarial
Graph Representation Adaptation (AGRA) framework that unifies graph representation propagation with adversarial learning for cross-domain holistic-local feature co-adaptation. \pxj{Different from the above works, our work utilizes AU information as auxiliary cues to bridge the gap of different FER datasets, which is expected to learn a generic feature space for source dataset and target dataset.}


\section{Methodology}
\label{method}
In this section, we first overview the AdaFER, and then present its key modules. We finally present the details of training and inference.


\subsection{Overview of AdaFER}
The goal of our method is to mitigate the domain gap including inconsistent annotations and different imaging conditions. Considering the relationship between subjective facial expression and objective Action Units, we propose a simple yet efficient AU-guided unsupervised Domain Adaptive Facial Expression Recognition (AdaFER) method.
Figure \ref{fig:pipeline} illustrates the pipeline of our AdaFER which mainly consists of two crucial modules: i) AU-Guided Annotating (AGA) and ii) AU-Guided Triplet Training (AGT).
Given images from source domain and target domain, we first utilize a pretrained AU detection model to extract the AUs coding for images from both domains.
Then the AGA module assigns a soft/hard pseudo label for each image in target domain by comparing the AUs between source and target domain. For example, one of the AGA strategies is to assign a target image with a hard label that is the same as a source image if both images have equal AUs. Meanwhile, we mine triplets among source and target domains according AUs. For example, given an anchor image in target domain, a positive image in source domain is the one with equal AUs and a negative image is the one with different AUs. We jointly train the FER model on source domain with ground-truths and target domain with pseudo labels and triplets.

\subsection{The AU Distributions of FER Datasets}
\begin{figure}[tp]
\center
\includegraphics[width=0.48\textwidth]{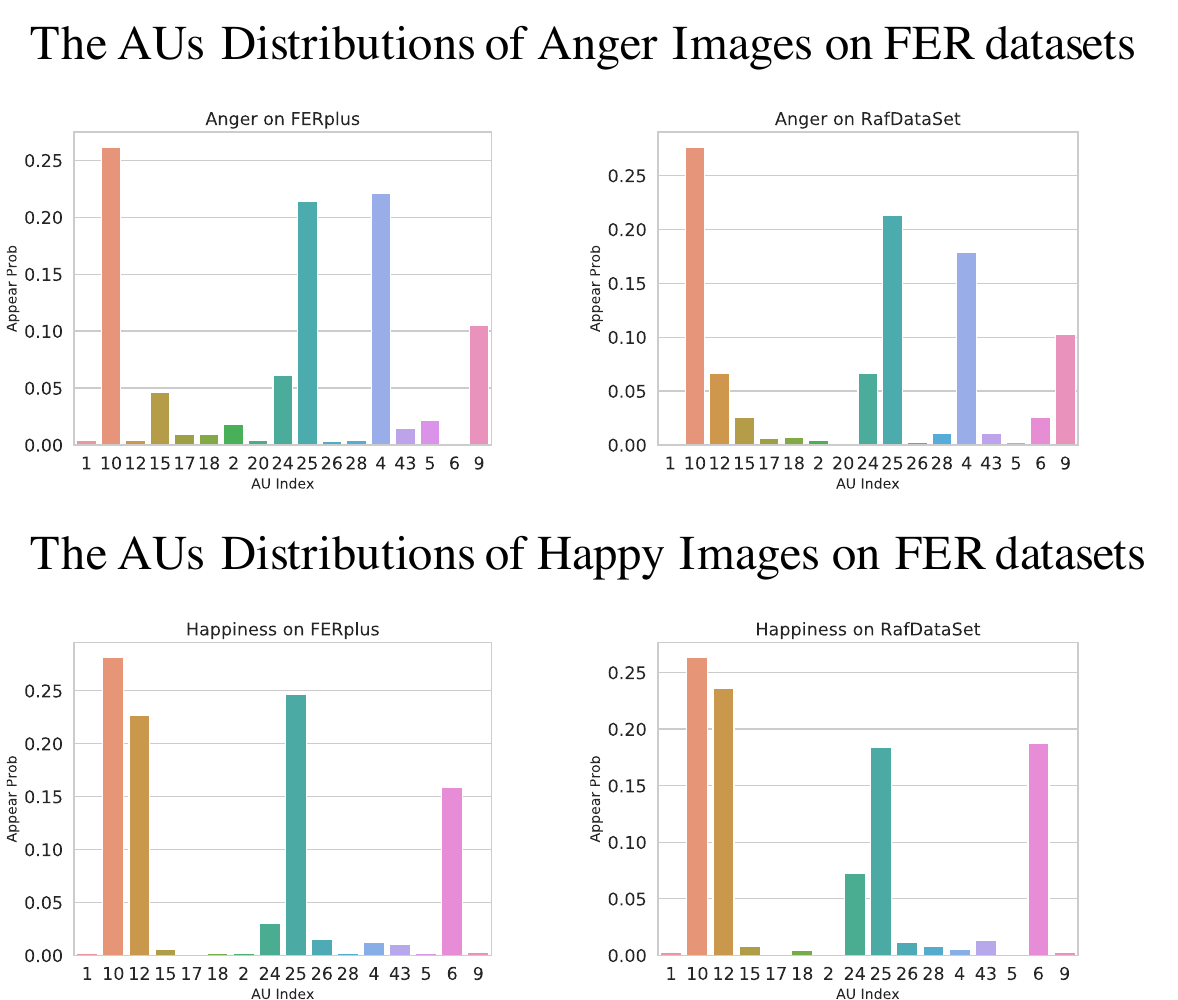}
	\caption{The distributions of AUs in different FER datasets.}
	\label{fig:au_dist}
\end{figure}

\kwang{To check whether AUs are with low bias among FER datasets (RAF-DB, FERPlus, ExpW, CK+), we visualize the AU distributions of each facial expression category. Specifically, we first utilize a pretrained AU detection model to extract AUs for each image. Then, we make statistics of AU occurrence numbers over each category. After normalizing, we show the AU statistics in Figure \ref{fig:au_dist}. We observe that i) the AU distributions of same categories are very similar among different datasets and ii) different facial expressions own very different AU distributions which indicates that AUs offer discriminate cues.}

\subsection{AU-Guided Annotating}
We introduce the AU-Guided Annotating (AGA) module to assign pseudo labels for target domain images according to AU detection results. Specifically, we elaborately design several assignment schemes as follows.

\textbf{Source-based hard label assignment} (S-hard).
Given an image $X_s^i$ and its detected AUs $E_s^i$ from source domain, the S-hard scheme utilizes $E_s^i$ as a query to search target domain images that have same AUs with $E_s^i$. This process can be defined as follows,

\begin{equation}
    [X_t^1,\ldots, X_t^k] = Q_{\rm{S-hard}}(E_s^i)=\{X_t^j~~ \textrm{if} ~~ E_t^j\equiv E_s^i \},
\end{equation}
where $[X_t^1,\ldots, X_t^k]$ are the retrieval images from target domain, $Q_{\rm{S-hard}}$ represents the S-hard operation. S-hard assigns all the retrieval images with the label of $X_s^i$.

\textbf{Target-based hard label assignment} (T-hard).
Different from S-hard, the T-hard scheme uses target domain images as query images. Given an image $X_t^i$ and its detected AUs $E_t^i$ from target domain, the query process can be formulated as follows,
\begin{equation}
    [X_s^1,\ldots, X_s^k] = Q_{\rm{T-hard}}(E_t^i)\{X_s^j~~ \textrm{if} ~~ E_s^j\equiv E_t^i \},
\end{equation}
where $[X_s^1,\ldots, X_s^k]$ represents the retrieval images from source domain. With the ground truths of source domain images, T-hard assigns the most frequent label to the target domain image $X_t^i$.

\textbf{Target-based soft label assignment} (T-soft).
For a query image from target domain, besides the hard assignment scheme, the T-soft directly uses the label distribution of retrieval samples to assign each target domain image a soft label vector. It is worth noting that there does not exist a source-based soft label assignment since we do not have the labels of target domain images.

\textbf{Learning with AGA.}
After assigning pseudo labels for target domain, we can train FER models in traditional ways. Suppose $Y_s \in \mathcal{R}^{{1}\times M}$ represents the labels of $M$ source domain images, $Y_{\rm{S-hard}} \in \mathcal{R}^{{1}\times N}$, $Y_{\rm{T-hard}} \in \mathcal{R}^{{1}\times N}$, $Y_{\rm{T-soft}} \in \mathcal{R}^{{C}\times N}$ denote the labels of $N$ target domain images in S-hard, T-hard, and T-Soft schemes, we use the following loss function by default to train with AGA module,
\begin{equation}
    L_c = \rm{CE}(P_s, Y_s) + \beta (\rm{CE}(P_t, Y_{\rm{S-hard}}) \\+  \rm{KL}(P_t, Y_{\rm{T-soft}})),
    \label{eq:AGAloss}
\end{equation}
where CE denotes the Cross-Entropy loss function, KL is the KL Divergence loss function, $P_t$ and $P_s$ represent the predictions of target and source domains images. $\beta$ is the trade-off ratio between the two loss values that calculate by pseudo labels and ground truths.

\subsection{AU-Guided Triplet Training}
To achieve structure-reliable and  compact facial features, we perform the AU-Guided Triplet Training (AGT) to further narrow the gap among different domains. The key step is to sample triplets from source and target domain. 

\textbf{Triplet Selection.} Intuitively, we can select triplets from the union of source domain and target domain. However, considering CD-FER is a classification task, we ignore the triplets in source domain since ground truths are available. Specifically, we only keep those cross-domain triplets.
 Given an anchor in source domain ($X_s^a$) or target domain ($X_t^a$), we first use it to retrieve the images of target domain or source domain that own the same AUs, and then we randomly select a positive sample from target domain ($X_t^p$) or source domain ($X_s^p$) according to the retrieval samples, and a negative sample $X_t^n$ or $X_s^n$ according to the rest samples of target domain or source domain. Thus, we mainly select two kinds of cross-domain triplets: ($X_s^a$,$X_t^p$,$X_t^n$) and ($X_t^a$,$X_s^p$,$X_s^n$). We conduct triplet selection in an offline manner. In addition, we also perform hard negative mining by sorting the similarities between the AU scores of anchor and all negative samples. We randomly select a sample from these with AU similarities larger than a threshold $\tau_n$ (0.5 by default) as a negative sample.
 


\textbf{AU-Guided Triplet Loss.} After the selection of triplets, we use triplet loss to learn discriminative and compact features as follows,
\begin{equation}
    L_{tri} = max\{0, \gamma - (\left\|F_a - F_n\right\|- \left\|F_a - F_p\right\|)\},
\end{equation}
where $F_a$, $F_p$, and $F_n$ represent the L2-normalized features of anchor, positive, and negative images, respectively. $\gamma$ is a margin which can be a fixed hyper parameter or a learnable parameter. We evaluate it in the experiment section. Training with these cross-domain triplets, we can obtain a cross-domain common feature space which makes similar facial images close and dissimilar ones far away. 
Considering both pseudo annotations and triplets, the total loss function is $L_{all} = L_{c} + \epsilon L_{tri}$ where $\epsilon$ is a trade-off ratio. 


\subsection{Implementation Details}
\textbf{AU detection and FER backbone}. Face images are detected and aligned by Retinaface \cite{deng2019retinaface} and further resized to 224 $\times$ 224 pixels. We utilize an advanced AU detector that pretrained on EmotiNet dataset using MLCR\cite{mlcr} algorithm, to extract AUs for each image. We then evaluate the effectiveness of AdaFER using ResNet-18 \cite{he2016deep} with Pytorch toolbox. The ResNet-18 is pre-trained on the MS-Celeb-1M face recognition dataset and the facial features for triplet training are extracted from the last pooling layer.

\textbf{Training}. We train our AdaFER in an end-to-end manner with 1 Tesla V100, and set the batch size as 128, \textit{i.e.}, 128 triplets with ground-truth labels or pseudo labels.  In each iteration, all the images are optimized by Cross-Entropy loss, KLDiv loss and AU-Guided Triplet Loss. The ratio $\beta$ and is defaulted as 1 and evaluated in the ensuing Experiments. The triplet loss margin $\gamma$ is set as 0.5 by default. The ratio of $L_{c}$ and $L_{tri}$ is empirically set at 1:1, and its influence will be studied in the ensuing ablation study of Experiments. The leaning rate is initialized as 0.001 with Adam optimizer using Exponential (gamma =0.9) scheme to reduce the learning rate. We stop training at 40-th epoch.

\section{Experiments}
\label{experiments}

In this section, we first describe employed datasets. We then demonstrate the robustness of our AdaFER in cross-domain facial expression recognition task. Further, we conduct ablation studies to show the effectiveness of each module and the settings of hyper-parameters in AdaFER. After that, We compare AdaFER to related state-of-the-art methods. Finally, to obtain a better understanding of AdaFER, we visualize the statistical distributions of CK+ and FERPlus datasets.

\subsection{Datasets}
\noindent \textbf{RAF-DB.}~\cite{li2017reliable} consists of 30,000 facial images annotated with seven basic and 14 compound facial expressions by 40 trained students. In our experiments, we default set RAF-DB as source dataset and only images with seven basic expressions (neutral, happiness, surprise, sadness, anger, disgust, fear, neutral) are used which leads to 12,271 images for training.

\noindent \textbf{FER2013 \& FERPlus}~\cite{BarsoumICMI2016} contain 28,709 training images, 3,589 validation images and 3,589 test images, all of which are resized to 48$\times$48 pixels. FER2013 is collected by Google Image Search API and annotated by seven facial expressions. However, the annotation of FER2013 is not accurate because the there are only two annotators. Therefore, FERPlus is extended from FER2013 as used in the \textit{ICML 2013 Challenges}, it is annotated by 10 annotators and added contempt category.

\noindent \textbf{CK+}~\cite{mollahosseini2017affectnet} contains 593 video sequences from 123 subjects.
Among these videos, 327 sequences from 118 subjects are labeled with seven expressions (except neutral), i.e. anger, contempt, disgust, fear, happiness, sadness, and surprise. All the subjects start from neutral and increase their expression intensity to seven expressions. Therefore, we select the last three frames with peak formation from each sequence and the first frame (neutral face) of each sequence, resulting in 1236 images. We follow previous work~\cite{mollahosseini2017affectnet} to choose 1108 images for training and 128 images for testing.

\noindent \textbf{ExpW}~\cite{zhang2018facial} contains 91793 images that are annotated by one of the seven expressions. Since the official ExpW dataset does not provide training/testing splits, we follow \cite{xie2020adversarial} to select 28848 for training, 28853 for validating and 28848 for testing.

\noindent \textbf{JAFFE}~\cite{lyons1998coding} collects 213 images from 10 Japanese females in lab-controlled condition. We here choose 170 images for training and 43 images for testing.

\begin{table}[t]
\center
\caption{Performance (\%) comparison between the proposed AdaFER and baseline methods
}
 \resizebox{\linewidth}{!}{
\begin{tabular}{@{}ccccccccccccccc@{}}
\toprule
Method  & CK+ & JAFFE  & ExpW & FER2013 & FERPlus \\
\hline
\#1 & 70.54 & 46.51  & 61.53 & 52.91 & 62.40\\
\#2 & 65.11 & 27.50 & 66.52 & 46.31 & 68.35 \\
\#3 &  74.42 & 55.81 & 68.13 & 55.89 & 63.81 \\
\#4 &  73.64 & 58.14 & 69.41 & 54.81 & 70.02 \\
\#5 &  71.32 & 53.49 & \textbf{73.58} & 57.15 & 77.99 \\
\textbf{AdaFER} &  \textbf{81.40} & \textbf{61.37}& 70.86 & \textbf{57.29}& \textbf{78.22}  \\
\bottomrule
\end{tabular}}
\label{tab:compare_baseline}
\end{table}

\subsection{AdaFER for Unsupervised CD-FER}
To evaluate the effectiveness of AdaFER, we compare several baseline methods with our proposed AdaFER using RAF-DB as source dataset and testing on CK+, JAFFE, ExpW, FER2013, and FERPlus.\pxj{We totally implement five baseline methods as follows.}
\begin{itemize}
    \item \#1: We train a model on source data and directly test on target data.
    \item \#2: We first extract AUs for both source and target data, and then use the AUs of each image in target data to query source data, and finally we assign the most frequent category of retrieval images to the target image.
    \item \#3: We first use the trained model on source data to predict hard pseudo labels of target data, then fine-tune the model on target data.
    \item \#4: It is identical with \#3 except for the predicted pseudo labels are kept as vectors (i.e. soft labels).
    \item \#5: We use both the image and detected AUs as inputs to train a classify on the source set and then fine-tune the model on pseudo soft labelled target data.
\end{itemize}

\pxj{The results are shown in Table \ref{tab:compare_baseline}. Several observations can be concluded as follows. First, our AdaFER almost outperforms  all other baseline methods with a large margin especially when testing on lab-controlled CK+ and JAFFE datasets. Second, using pseudo labels (\#3 and \#4) to fine-tune model on target data also achieves large improvement over \#1. Third, the naive AU-based method (\#2) performs better than \#1 on FERPlus and ExpW which indicates AUs are useful among similar data. Last but not the least, the naive AU-based method degrades in lab-controlled FER datasets which suggests there exists AU domain gap between in-the-wild datasets and lab-controlled datasets. Nevertheless, our AdaFER mitigates the domain gap and achieves large improvements on both in-the-wild and lab-controlled datasets}

\begin{figure}[t]
	\includegraphics[width=0.48\textwidth]{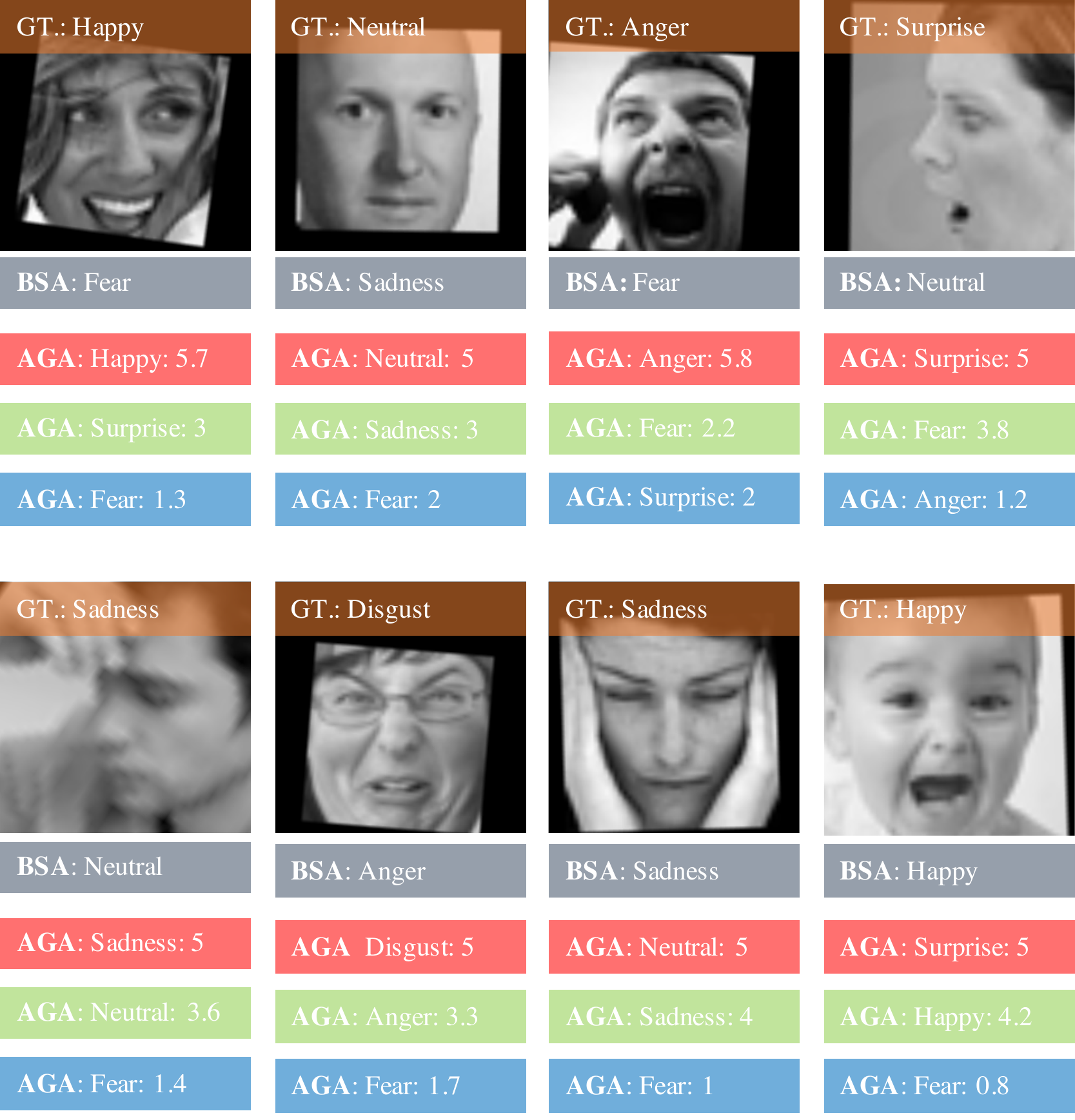}
	\caption{Visualization of pseudo labels from baseline \#3 and our AGA module. BSA represents baseline \#3.}
	\label{fig:au-query}
\end{figure}

\textbf{Visualization of AU-Guided Annotating.}
\pxj{To further investigate AdaFER, we visualize the pseudo labels of target images annotated by AGA module and baseline \#3. We use our T-soft assignment and list the top-3 categories. As shown in Fig \ref{fig:au-query}, AGA achieves better results than baseline \#3. In the first 6 images, AGA assigns the highest weights on the ground-truth category. The last two bad cases are  extremely ambiguous even for human, and our AGA seems to assign reasonable categories for them. }

\begin{table}[t]
\center
\caption{Evaluation the influences of anchor images for AGA module.}
\begin{tabular}{@{}cccccccccccccccc@{}}
\toprule
Source & Target & CK+  & ExpW & FER2013 & FERPlus \\
\hline
$\surd$ & $\times$ & 80.62  & 69.90 & 54.36 & 77.61  \\
$\times$ & $\surd$ & 71.23 & 69.73 & 55.45 & 77.93 \\
$\surd$ & $\surd$ & \textbf{81.40} & \textbf{70.86} & \textbf{57.29} & \textbf{78.22} \\
\bottomrule
\end{tabular}
\label{tab:evaluation_anchor_source}
\end{table}

\subsection{Ablation Studies}
We conduct ablation studies for the modules of our AdaFER and other hyper parameters.
\begin{figure}[tp]
\center
\includegraphics[width=0.48\textwidth]{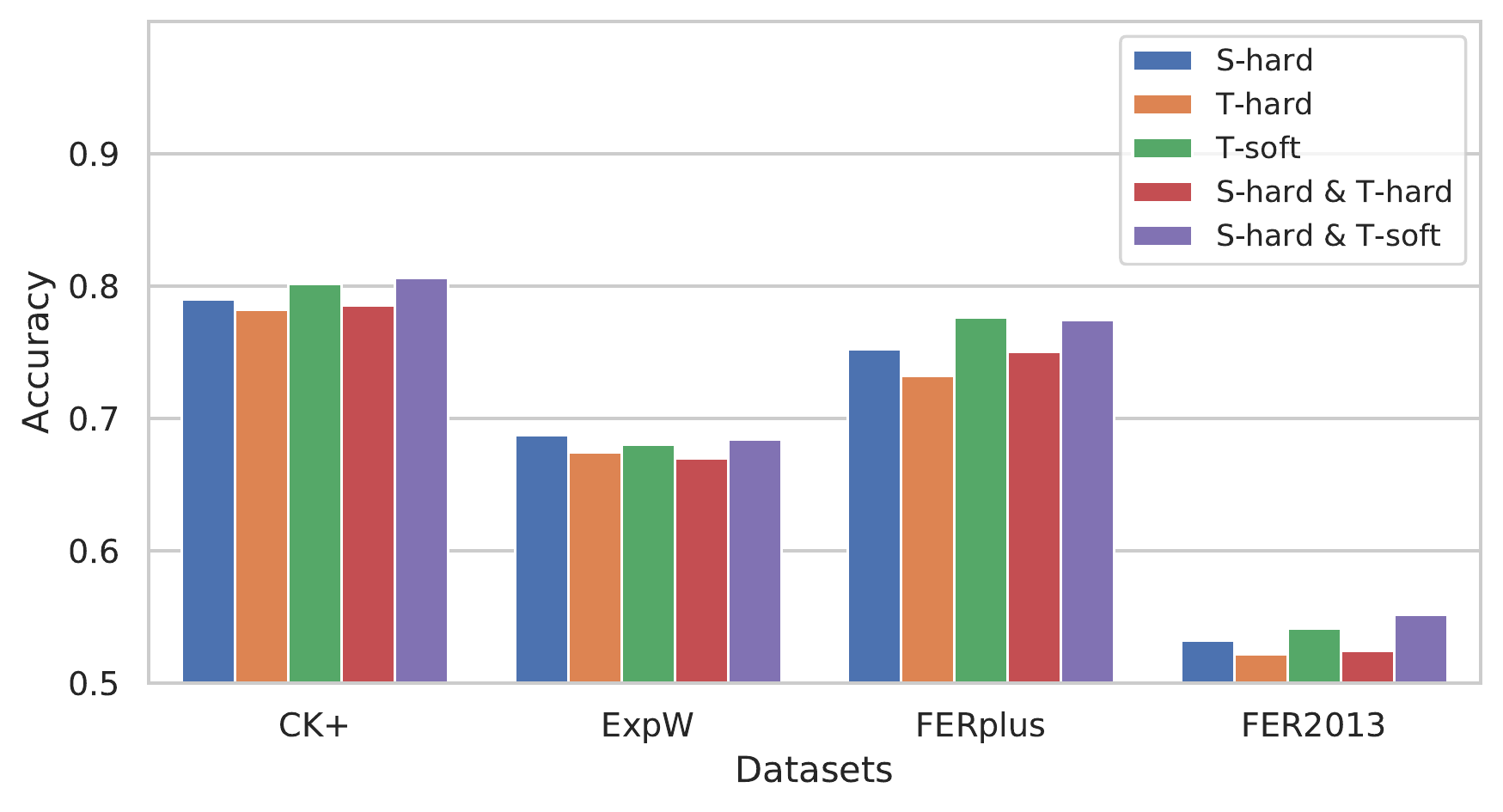}

	\caption{Evaluation of pseudo assignment strategies .}
	\label{fig:assignment}
\end{figure}

\begin{table}[t]
\center
\caption{Evaluation of AU-Guided Annotating (AGA) and AU-Guided Triplet Training (AGT) on CK+, ExpW, FER2013, and FERPlus.  }
\resizebox{\linewidth}{!}{
\begin{tabular}{@{}cccccccccccccccc@{}}
\toprule
AGA & AGT & CK+  & ExpW & FER2013 & FERPlus \\
\hline
$\times$ & $\times$ & 70.54 & 61.53 & 52.91 & 62.40\\
$\surd$ & $\times$ & 80.62 & 68.45 & 55.20  & 77.42 \\
$\times$ & $\surd$ & 80.28 & 69.80 & 56.45 & 74.50 \\
$\surd$ & $\surd$ & \textbf{81.40} & \textbf{70.86} & \textbf{57.29}& \textbf{78.22}  \\
\bottomrule
\end{tabular}}
\label{tab:evaluation_each_module}
\end{table}


\pxj{
\textbf{The three types of label assignments.}
In AGA, we introduce three kinds of assignment, namely S-hard, T-hard, and T-soft. We explore individual types and their combination on CK+, ExpW, FERPlus, and FER2013. The results are shown in Figure \ref{fig:assignment}.
For individual assignments, we observe that T-soft strategy performs best in average following by S-hard strategy. Combining T-soft and S-hard strategies further boost performance in most cases. We use this combination strategy by default in the following experiments. }

\pxj{
\textbf{Anchor images in AGT module.}
In AU-guided triplet training, both source and target data can be used as anchor images. We evaluate the effect of anchor images in Table \ref{tab:evaluation_anchor_source}. As can be observed, using source data as anchor images largely outperforms the target anchor scheme on CK+ which may be explained by that the target CK+ dataset is too small to collect enough triplet samples. Both individual anchor schemes perform similarly on large-scale datasets, and combining both schemes consistently boosts performance on all datasets.}

\begin{figure}[tp]
\center
\includegraphics[width=0.48\textwidth]{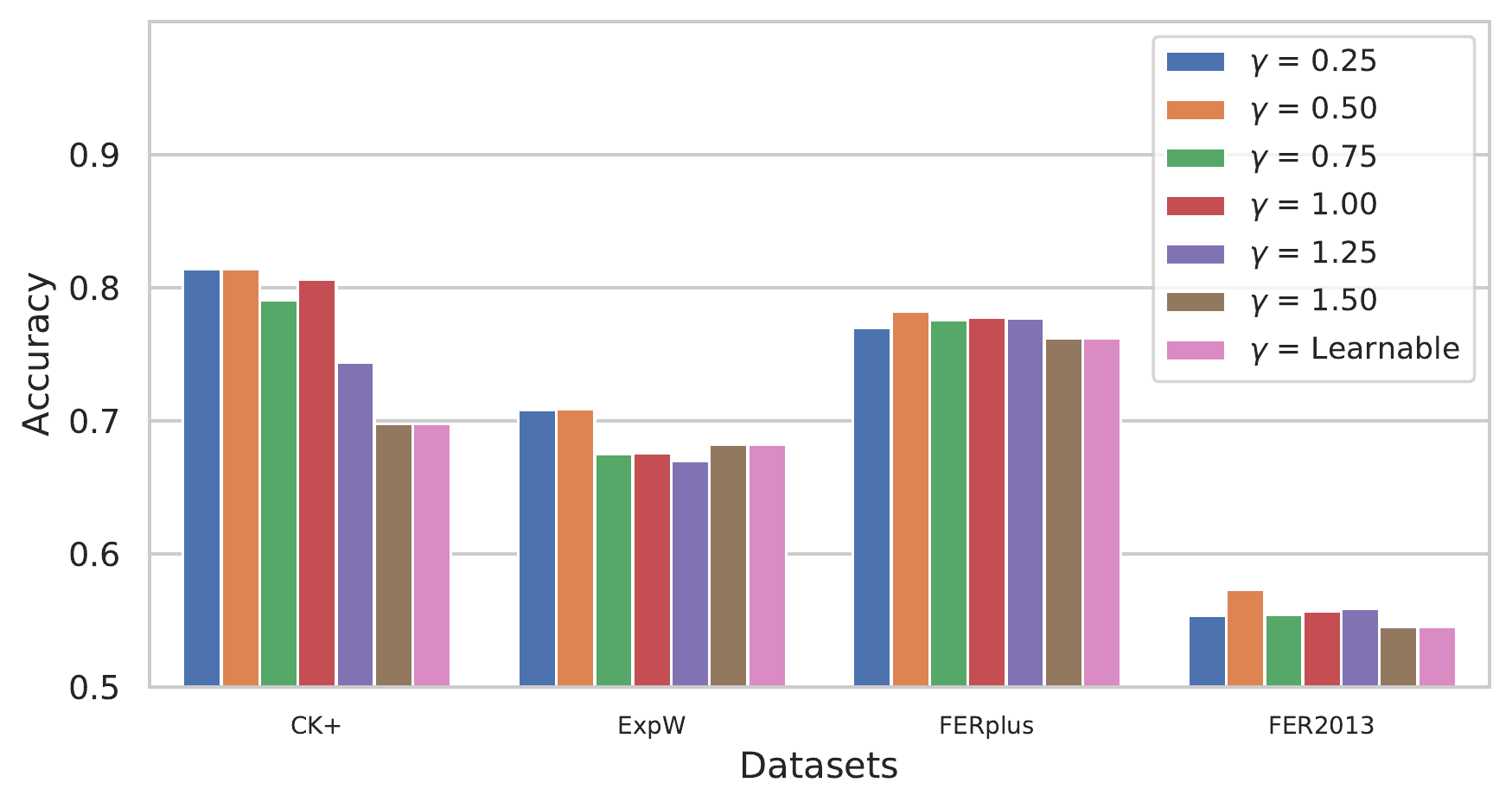}

	\caption{Evaluation of the margin parameter in AU-Guided Triplet Loss.}
	\label{fig:margin}
\end{figure}

\begin{figure}[tp]
\center
\includegraphics[width=0.5\textwidth,height=0.35\textwidth]{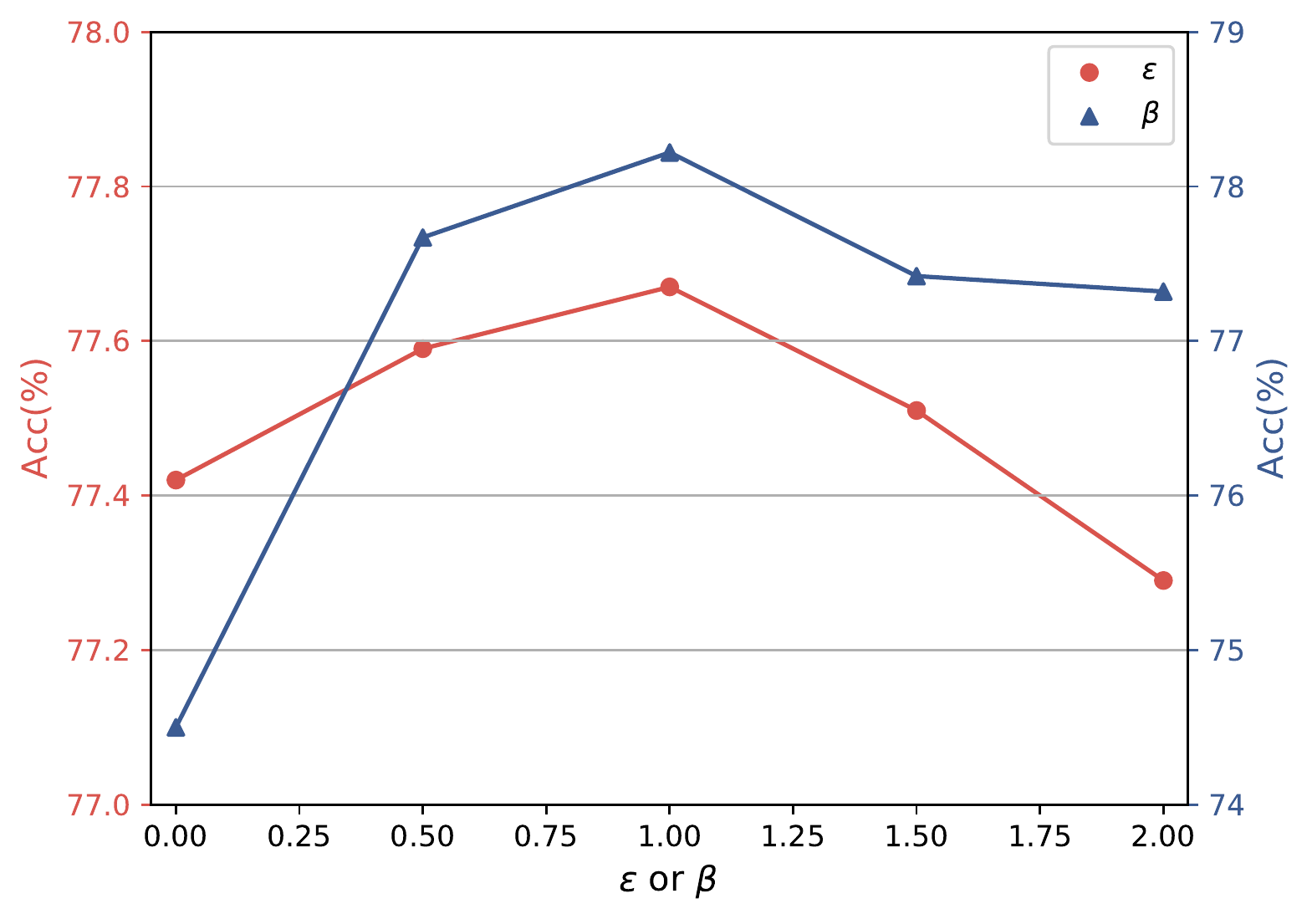}

	\caption{Evaluation of the trade-off ratios $\epsilon$ and $\beta$.}
	\label{fig:epsilon}
\end{figure}

\pxj{
\textbf{AU-Guided Annotating (AGA) and AU-Guided Triplet Training (AGT)} are two crucial modules in AdaFER which respectively leverages pseudo labels for target domain images and constraints the distance structure of each triplet tuple. To explore the effectiveness of each module, we conduct the evaluation on CK+, ExpW, FER2013, and FERPlus. As shown in Table \ref{tab:evaluation_each_module}, both AGA and AGT can individually improve the baseline by a large margin, and they perform similarly on most of datasets. This may be explained by that both AGA and AGT essentially resorts to AU information, and the only difference is that AGA performs supervision on classifier while AGT on feature mapping. Nevertheless, from the results of AGA+AGT, they are complementary with each other on all the datasets.}



\pxj{
\textbf{The margin $\gamma$ of AGT.} $\gamma$ is a margin parameter to control the distance between anchor-positive pair and anchor-negative pair. Theoretically, it can be a learnable parameter in the end-to-end framework. We evaluate it with both fixed mode and learnable mode. The results are illustrated in Figure \ref{fig:margin}. For the fixed mode, we evaluate margin from 0.25 to 1.5 with 0.25 as the interval. On all the 4 FER datasets, our default margin $\gamma = 0.5$ achieves the highest performance. Larger margin makes training harder which degrades performance on CK+ and ExpW. For the learnable mode, $\gamma$ respectively converges to 0.62 ($\pm 0.034$), 0.37 ($\pm 0.067$), 0.71($\pm 0.026$), and 0.42($\pm 0.033$) on CK+, ExpW, FERPlus, and FER2013. Meanwhile, the learnable $\gamma$ also obtains competitive results. 
}

\textbf{The trade-off ratios $\beta$ and $\epsilon$.}
$\beta$ is the trade-off ratio between two loss values calculated by pseudo labels and ground truths in Equation (\ref{eq:AGAloss}). $\epsilon$ is the trade-off ratio between $L_c$ and $L_{tri}$. We evaluate them from 0.0 to 2.0 with 0.25 as the interval on FERPlus dataset and present the results in Figure \ref{fig:epsilon} . For both $\beta$ and $\epsilon$, the final performance increases gradually and reaches the peak in value 1.0. Larger values degrade performance dramatically which illustrates that all the loss items are almost equally important.

\begin{table}[t]
\center
\caption{Evaluation of the threshold for hard negative mining.}
\begin{tabular}{@{}cccccccccccccccc@{}}
\toprule
Threshold & CK+  & ExpW & FER2013 & FERPlus \\
\hline
0 & 68.99  & 65.24 & 52.21     & 66.68 \\
0.25 & 72.03 & 68.23 & 56.23 & 74.23  \\
0.5  & \textbf{81.40} & \textbf{70.86} & \textbf{57.29} & \textbf{78.22}   \\
0.75 & 80.96 & 69.12 & 55.80 & 77.72  \\
\bottomrule
\end{tabular}
\label{tab:threshold}
\end{table}

\begin{table*}[htp]
\center
\caption{Comparison to the state-of-the-art methods on CK+, JAFFE, FER2013, ExpW datasets. The results of upper part are taken from the corresponding papers, and the results of the bottom part are taken from a cross-domain facial expression recognition benchmark's \cite{tianshui2020adversarial} implementations. The last column shows the mean accuracy of performances on all the datasets.}
\resizebox{\linewidth}{!}{
\begin{tabular}{@{}cccccccccccccccc@{}}
\toprule
Methods &Source Dataset & Backbones & CK+ &JAFFE  & FER2013 & ExpW & Mean \\
\hline
 Da et al.\cite{da2015effects}& BOSPHORUS &HOG \& Gabor Filters & 57.60 &36.2  &- &- &-  \\
 Hasani et al.\cite{hasani2017facial} &MMI\&FERA\&DISFA &Inception-ResNet &67.52 &-  &- &- &-  \\
 Hasani et al.\cite{hasani2017spatio}&MMI\&FERA &Inception-ResNet &73.91 &-  &- &- &- \\
 Zavarez et al.\cite{zavarez2017cross}&Six Datasets &VGG-Net & 88.58 &44.32  &- &- &-  \\
 Mollahosseini et al.\cite{mollahosseini2016going}&Six Datasets &Inception &64.20 &-  &34.00 &- &-  \\
  DETN\cite{li2020deep}&RAF-DB &Manually-Designed Net & 78.83 & 57.75 & 52.37 &- &-  \\
 ECAN\cite{li2020deeper}&RAF-DB 2.0 &VGG-Net &86.49 &61.94  &58.21 &- &-  \\
 \hline
 CADA\cite{long2018conditional}&RAF-DB &ResNet-18 &73.64 &55.40  &54.71 &63.74 & 61.87  \\
 SAFN\cite{xu2019larger}&RAF-DB &ResNet-18 & 68.99 & 49.30 & 53.31 & 68.32 & 59.98  \\
 SWD\cite{lee2019sliced}&RAF-DB &ResNet-18 & 72.09 & 53.52 & 53.70 & 65.85 & 61.29  \\
 LPL\cite{li2017reliable}&RAF-DB &ResNet-18 & 72.87 & 53.99 & 53.61 & 68.35 &62.20  \\
 DETN\cite{li2020deep}&RAF-DB &ResNet-18 & 64.19 &52.11 &42.01 &43.92 &50.55 \\
 ECAN\cite{li2020deeper}&RAF-DB &ResNet-18 &66.51 &52.11 &50.76 &48.73 &54.52  \\
 AGRA\cite{xie2020adversarial}&RAF-DB &ResNet-18 &77.52 &61.03 &54.94 &69.70 &65.79  \\
 \textbf{AdaFER}&RAF-DB &ResNet-18 & \textbf{81.40} & \textbf{61.37}& \textbf{57.29} & \textbf{70.86}& \textbf{67.73}  \\
 Baseline (\#3) &FERPlus &ResNet-18 & 64.34 & 41.86 & - & 66.64 & 57.87  \\
 \textbf{AdaFER}&FERPlus &ResNet-18 & 65.12 & 46.51& - & 73.58 & 61.47  \\
 
\bottomrule
\end{tabular}}
\label{tab:sota}
\end{table*}


\pxj{
\textbf{The threshold for hard negative mining of triplet samples.} In triplet selection, we sort the AU scores and set a threshold for negative mining. We evaluate the threshold from 0 to 0.75 in Table \ref{tab:threshold}. On all datasets, increasing the threshold from 0 to 0.5 boosts performance largely. Too small threshold could make zero triplet loss since triplet samples are too easy. Too large threshold may introduce hard positive samples as negative ones which is harmful for training.}

\begin{figure}[tp]
\center
\includegraphics[width=0.48\textwidth]{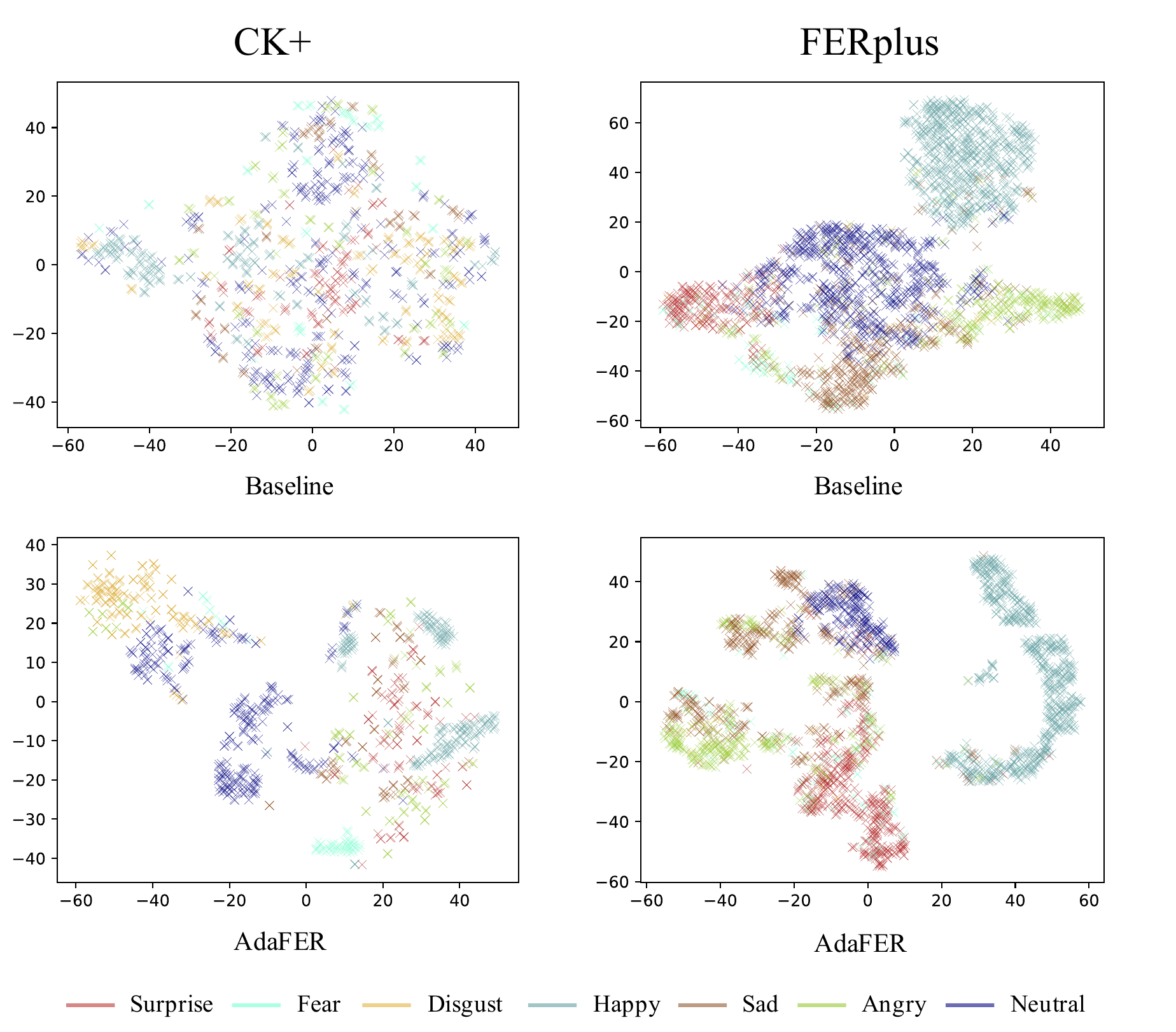}

	\caption{Visualization of feature distributions on CK+ and FERPlus datasets. The top and bottom parts show the feature distributions using baseline method and AdaFER, respectively.}
	\label{fig:distribution}
\end{figure}

\subsection{Comparison to state-of-the-art methods}

Table \ref{tab:sota} compares our AdaFER with 13 state-of-the-art (SOTA) methods in CD-FER task. The methods from the bottom table use RAF-DB as source dataset, and these from the upper table use other datasets. As shown in Table \ref{tab:sota}, our AdaFER achieves competitive results compared to SOTA methods. Zavarez \textit{et al.}\cite{zavarez2017cross} achieve the SOTA on CK+ using six datasets. ECAN\cite{li2020deeper} obtain SOTA on JAFFE and FER2013 using a model pretrained on VGGFace2 and then fine-tune on RAF-DB2.0 which is not publicly available. 
For fair comparison, the methods in the bottom part of Table \ref{tab:sota} use the same source dataset and backbone. The mean accuracy over all target datasets is also computed for easy comparison. Our AdaFER obtains accuracy \textbf{81.40\%}, \textbf{61.37\%}, \textbf{57.29\%}, and \textbf{70.86\%} on CK+, JAFFE, FER2013, and ExpW datasets, which are the new state-of-the-art CD-FER results on these datasets. Moreover, our AdaFER does not increase any computing cost in inference phase. AGRN \cite{xie2020adversarial} needs to extract holistic and local features to initialize the nodes of target domain in inference stage, which is time-cosuming. LPL \cite{li2017reliable} relies on the assumption of both source and target domains. 

\pxj{To evaluate the robustness of our method on different source datasets, we also show the CD-FER performance with FERPlus as source dataset. Note that the FER2013 dataset has the same images as FERPlus, therefore we ignore the performance on FER2013. Our AdaFER also improve the baseline (\#3) with large margin in term of mean accuracy.}

\pxj{
\textbf{Understanding AU-Guided Learning}.
To better understand the differences between baseline (\#3) and our AdaFER, we utilize the T-SNE method to illustrate the feature distributions of the test sets of CK+ and FERPlus. The results are shown in Figure \ref{fig:distribution}. We can find that our AdaFER can learn more compact features than baseline in unsupervised CD-FER tasks. For CK+, `Neutral' samples are the most scattered ones for the baseline method, AdaFER can cluster them dramatically. For FERPlus, the samples of same categories are clustered compactly while these of different categories are largely separated. These illustrates the reason why AdaFER improves baseline dramatically.}

\section{Conslusion}
\label{conclusion}
In this paper, we address the unsupervised domain adaptive facial expression recognition task with the auxiliary facial action units. Our method is very different from existing cross-domain FER methods which typically follow generic cross-domain methods. Specifically, we proposed an AU-guided unsupervised Domain Adaptive FER (AdaFER) framework which includes an AU-guided annotating module and an AU-guided triplet training module. We evaluated several AU-guided annotating strategies and triplet selection methods. Extensive experiments on several popular benchmarks show that i) AdaFER achieves state-of-the-art results and ii) obtains discriminative and compact facial expression features.


{\small
\bibliographystyle{ieee_fullname}
\bibliography{references_fer}
}

\end{document}